\PassOptionsToPackage{doi=false,isbn=false,url=false}{bibtex}
\documentclass[letterpaper, 10 pt, conference]{ieeeconf} 

\IEEEoverridecommandlockouts                              

\usepackage{graphicx} 
\graphicspath{ {./figures/} }
\usepackage{amsmath} 
\usepackage{amssymb}
\usepackage{algorithm}
\usepackage[noend]{algpseudocode}
\usepackage{xcolor}
\definecolor{tumblue}{RGB}{0,101,189}

\usepackage{stfloats}
\usepackage{booktabs}
\usepackage{setspace}
\usepackage{tabularx}
\usepackage{booktabs}
\usepackage[T1]{fontenc}
\usepackage[scaled=0.85]{beramono}
\usepackage{listings}
\usepackage{caption}
\usepackage{cleveref}
\usepackage{url}
\let\labelindent\relax
\usepackage{enumitem}

\Crefname{lstlisting}{Listing}{Listings}
\lstdefinestyle{sparqlstyle}{
  language=OCL,
  basicstyle=\footnotesize,
  stepnumber=1,
  numbersep=10pt,
  tabsize=2,
  showspaces=false,
  breaklines=true
}

\usepackage[nonumberlist,nopostdot,acronym,toc]{glossaries}
\newacronym{cca}{CCA}{Central Controller Agent}
\newacronym{llm}{LLM}{Large Language Model}
\newacronym{mas}{MAS}{Multi-Agent System}
\newacronym{pa}{PA}{Product Agent}
\newacronym{ra}{RA}{Resource Agent}
\newacronym{api}{API}{Application Programming Interface}
\newacronym{sme}{SME}{Subject Matter Expert}
\newacronym{fsm}{FSM}{Finite State Machine}
\newacronym{RepastS}{RepastS}{Repast Symphony}
\newacronym{CoT}{CoT}{Chain-of-Thought}
\newacronym{mtbf}{MBTF}{Mean Time Between Failures}
\newacronym{mttr}{MTTR}{Mean Time To Repair}
\newacronym{rag}{RAG}{Retrieval Augmented Generation}
\newacronym{agv}{AGV}{Automated Guided Vehicle}

\usepackage{balance} 
\usepackage{placeins}

\makeatletter
\newcommand\fs@betterruled{%
  \def\@fs@cfont{\bfseries}\let\@fs@capt\floatc@ruled
  \def\@fs@pre{\vspace*{5pt}\hrule height.8pt depth0pt \kern2pt}%
  \def\@fs@post{\kern2pt\hrule\relax}%
  \def\@fs@mid{\kern2pt\hrule\kern2pt}%
  \let\@fs@iftopcapt\iftrue}
\floatstyle{betterruled}
\restylefloat{algorithm}
\makeatother

\author{Jonghan Lim$^{1}$ and Ilya Kovalenko$^{2}$
\thanks{$^{1}$Jonghan Lim is with the Department of Industrial and Manufacturing, Pennsylvania State University, State College, USA
        (e-mail: jxl567@psu.edu).
        }%
\thanks{$^{2}$ Ilya Kovalenko is with the Department of Industrial and Manufacturing and the Department of Mechanical Engineering, Pennsylvania State University, State College, USA
        (e-mail: iqk5135@psu.edu).
        }%
}

\title{\LARGE \bf
Dynamic Task Adaptation for Multi-Robot Manufacturing Systems with Large Language Models
}

\begin{document}
\setlength{\textfloatsep}{2pt}

\maketitle
\thispagestyle{empty}
\pagestyle{empty}

\begin{abstract}
 
Recent manufacturing systems are increasingly adopting multi-robot collaboration to handle complex and dynamic environments. While multi-agent architectures support decentralized coordination among robot agents, they often face challenges in enabling real-time adaptability for unexpected disruptions without predefined rules. Recent advances in large language models offer new opportunities for context-aware decision-making to enable adaptive responses to unexpected changes. This paper presents an initial exploratory implementation of a large language model-enabled control framework for dynamic task reassignment in multi-robot manufacturing systems. A central controller agent leverages the large language model's ability to interpret structured robot configuration data and generate valid reassignments in response to robot failures. Experiments in a real-world setup demonstrate high task success rates in recovering from failures, highlighting the potential of this approach to improve adaptability in multi-robot manufacturing systems.

\end{abstract}

\section{Introduction}
\label{sec:introduction}

Manufacturing processes demand high precision, speed, and adaptability to accommodate shorter product lifecycles for growing customization needs. Robotic systems address these demands by providing consistent precision, speed, and reliability, making them effective for high-throughput and repetitive tasks. Building on these strengths, manufacturers are progressively adopting multi-robot collaboration to improve operational efficiency, flexibility, and scalability on the shop floor, such as assembly lines~\cite{marvel2018multi}, and large-scale additive manufacturing~\cite{shen2019research}. Coordinated task execution allows parallel processing, workload allocation, and enhanced adaptability across complex and dynamic production scenarios~\cite{perez2020digital}. Despite these advantages, reconfiguring and adapting multi-robot systems remains a complex challenge. They often require a significant effort to manage coordination, task allocation, and scalability in dynamic industrial settings.

To enhance adaptability in dynamic environments,~\gls{mas} architectures have been studied to coordinate resources and tasks~\cite{leitao2009agent}. By decentralizing decision-making through autonomous agents that represent individual robots, machines, and products,~\gls{mas} enables more flexible and responsive control in manufacturing systems~\cite{leitao2012past}.~\gls{mas} approaches have also been applied to coordinate multi-robot cooperative additive manufacturing, enabling decentralized decision-making and improved robustness under dynamic conditions~\cite{poudel2023decentralized}. While~\gls{mas}-based multi-robot systems enable decentralized coordination and task allocation, they often struggle to adapt effectively when unexpected disruptions occur due to predefined rules and limited real-time reasoning capabilities.

Due to the emergence of~\glspl{llm}, they have been introduced to multi-robot domains to improve task planning, coordination, and adaptability in complex environments. Recent frameworks such as SMART-LLM~\cite{kannan2024smart} and DART-LLM~\cite{wang2024dart} leverage the reasoning and generalization capabilities of~\glspl{llm} to perform high-level task decomposition and robot-specific task allocation. These approaches enable robots to interpret ambiguous natural language instructions and execute tasks with minimal hard-coded logic. By incorporating mechanisms, such as few-shot prompting, hybrid communication, and dependency-aware planning,~\gls{llm}-enabled systems demonstrate improved scalability and responsiveness in both simulated and real-world multi-robot scenarios.

Despite recent progress, significant challenges remain in applying~\glspl{llm} to real-world multi-robot manufacturing systems. A key challenge is the lack of real-time adaptability to dynamic disruptions such as robot failures. Most existing~\gls{llm}-based approaches focus on task decomposition and planning, with limited focus on dynamic conditions of current manufacturing environments. Moreover, effectively utilizing~\glspl{llm} to adapt to heterogeneous robot platforms requires further investigation into how they can interpret structured representations of robot capabilities and environmental constraints. To address these challenges, \glspl{llm} can be integrated into the central controller architecture proposed by Kovalenko et al.~\cite{kovalenko2022toward} for~\gls{mas}, enabling enhanced adaptability and dynamic capability exploration in response to unexpected disruptions.

This paper presents an initial exploratory implementation of an~\gls{llm}-enabled control framework for dynamic task adaptation in multi-robot manufacturing systems. The key contributions of this work include:
(1) designing a preliminary~\gls{cca} that utilizes~\gls{llm} reasoning to reassign tasks under robot disruptions,
(2) proposing a structured prompting and feedback method that guides the~\gls{llm} to generate valid configuration data under real-world physical constraints, and
(3) conducting a proof-of-concept demonstration on a physical dual-robot setup to assess the ability to recover from disruptions through context-aware task reassignment.

The remainder of this paper is organized as follows.
Section~\ref{sec:problem} formulates the problem of dynamic task adaptation in multi-robot manufacturing systems.
Section~\ref{sec:system} presents the overall system architecture and implementation details of the proposed~\gls{llm}-enabled control framework.
Section~\ref{sec:experiments} reports experimental results from a physical testbed evaluating the system's performance under robot disruption scenarios.
Finally, Section~\ref{sec:conclusion} concludes the paper and outlines potential directions for future work.

\section{Problem Description}
\label{sec:problem}

This section describes the main components of the dynamic task adaptation problem in a multi-robot manufacturing system. Specifically, we define the robot set, capability configurations, task constraints, and the condition under which task reassignment is valid when a robot becomes disrupted.

The set of all robots are defined as: $\mathcal{R} = \{ r_1, r_2, \dots, r_N \}$.
Each robot \( r_i \in \mathcal{R} \) is characterized by a capability configuration: $
\mathcal{K}_{r_i} = \{ (k_1, v_1), (k_2, v_2), \dots, (k_m, v_m) \}$, where each \( k_j \) represents a capability key (e.g., reachability, sensing) and \( v_j \) is its corresponding value. The set of tasks for robots: $ \mathcal{T} = \{ \tau_1, \tau_2, \dots, \tau_M \}
$
Each task \( \tau_j \in \mathcal{T} \) is associated with a set of constraints:
$
\mathcal{C}_{\tau_j} = \{ \varphi_1, \varphi_2, \dots, \varphi_K \}
$
where each \( \varphi_l : \mathcal{K}_r \rightarrow \{ \text{valid}, \text{invalid} \} \) is a boolean-valued function that encodes a requirement for task execution.

Suppose \( r_d \in \mathcal{R} \) is the disrupted robot originally assigned to task \( \tau_d \in \mathcal{T} \). The system identifies another robot \( r_e \in \mathcal{R} \), referred to as the exploration robot, and generates an updated configuration \( \mathcal{K}_{r_e}' \). The reassignment is valid if the updated configuration satisfies all constraints associated with the disrupted task:
\[
\forall \varphi_l \in \mathcal{C}_{\tau_d}, \quad \varphi_l(\mathcal{K}_{r_e}') = \text{valid}
\]
Each constraint function \( \varphi_l \) encodes a requirement, such as spatial reachability, sensory compatibility, or tool capability, and returns \texttt{valid} if the robot's updated configuration \( \mathcal{K}_{r_e}' \) fulfills the condition.
For example, if the drop-off location of a task falls within the defined reachability limits of a robot, the spatial constraint is valid and the task can be reassigned. This formulation guarantees that the exploration robot, $r_e$, after reconfiguration, meets all requirements of task \( \tau_d \) and can safely resume its execution.

\section{System Overview and Implementation}
\label{sec:system}

\subsection{System Overview}
\label{subsec:system}

\begin{figure}[t]
\smallskip
\smallskip
    \captionsetup{belowskip=-1pt}    \includegraphics[width=.48\textwidth]{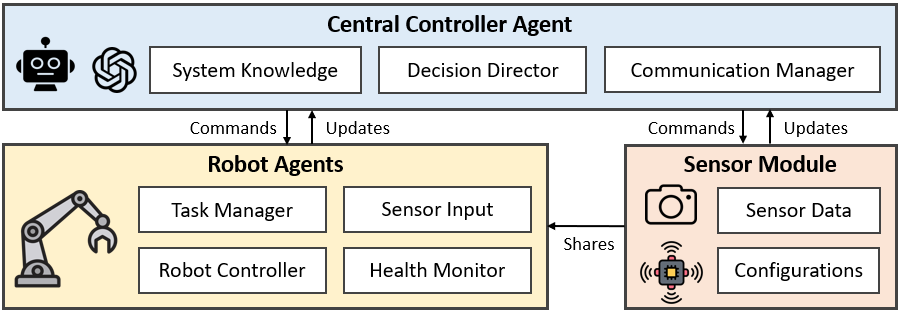}
    \caption{An~\gls{llm}-enabled ~\gls{cca} Framework for Multi-Robot Manufacturing Systems}
    \label{fig:overview}
\end{figure}

This section presents the overall architecture of the \gls{llm}-enabled~\gls{cca} framework for multi-robot manufacturing systems, as illustrated in Figure~\ref{fig:overview}. The system has three main components: the~\gls{cca}, robot agents, and a sensor module.

\subsubsection{Central Controller Agent}
\label{subsec:cca}

The~\gls{cca} serves as the main controller for enabling dynamic task adaptation in a multi-robot manufacturing system. The~\gls{cca} operates over the set of robots \( \mathcal{R} \), tasks \( \mathcal{T} \), and capability configurations \( \{ \mathcal{K}_{r_i} \}_{r_i \in \mathcal{R}} \). Upon disruption of a robot \( r_d\), originally assigned to a task \( \tau_d \in \mathcal{T} \), the~\gls{cca} searches for a candidate robot \( r_e \in \mathcal{R} \) and generates an updated configuration \( \mathcal{K}'_{r_e} \) by utilizing the reasoning capabilities of the~\gls{llm}. To enable context-aware dynamic adaptation, the~\gls{cca} maintains \textit{System Knowledge} on the manufacturing environment, including robot configurations, buffer positions, and sensor mappings. Based on this system knowledge, the~\gls{cca} acts as the \textit{Decision Director} that handles robot failures, utilizing the~\gls{llm}, and validating the resulting configurations based on task constraints. The~\gls{cca} also includes a \textit{Communication Manager} that applies updated robot configurations through a refresh configuration function, ensuring that new tasks and sensor assignments are integrated into the system. These components within the~\gls{cca} reason over disruptions, validate reassignments, and reconfigure them in real time.

\subsubsection{Robot Agents}
\label{subsec:robot-agent}

Each robot agent \( r_i \in \mathcal{R} \) is responsible for executing assigned tasks, using real-time perception and configuration-defined behaviors. The \textit{Task Manager} maintains the set of executable tasks \( \mathcal{T} \) assigned to the robot by the user. The \textit{Sensor Input} collects real-time perceptual data, coming from the sensor module. The \textit{Robot Controller} transforms perception-driven inputs into executable motion sequences. Given a pickup or drop-off task associated with an object, the robot executes movement using parameters in \( \mathcal{K}_{r_i} \) (e.g., speed, location). The \textit{Health Monitor} passively monitors operational status. In the event of a disruption, the robot transitions to a failure state and notifies the~\gls{cca}. 

\subsubsection{Sensor Module}
\label{subsec:sensor-module}

The sensor module manages real-time perception for the multi-robot manufacturing system, enabling environment-driven decision-making and adaptive task execution. Let \( \mathcal{O} = \{ o_1, o_2, \dots, o_L \} \) represent the set of perceptual observations, and let \( \mathcal{S}_{r_i} \) denote the structured representation of sensor data available to robot \( r_i \in \mathcal{R} \). These observations are interpreted in the robot’s local frame and used for task executions. Each task \( \tau_j \in \mathcal{T} \) is associated with a specific set of observability requirements $\mathcal{V}_{\tau_j} \subseteq \mathcal{S}$, indicating which sensory modalities or data sources are required to perceive the task-relevant context (e.g., object location, orientation).

\subsection{Implementation}
\label{subsec:system}

\begin{figure}[t]
\smallskip
\smallskip
    \captionsetup{belowskip=5pt}
    \includegraphics[width=.48\textwidth]{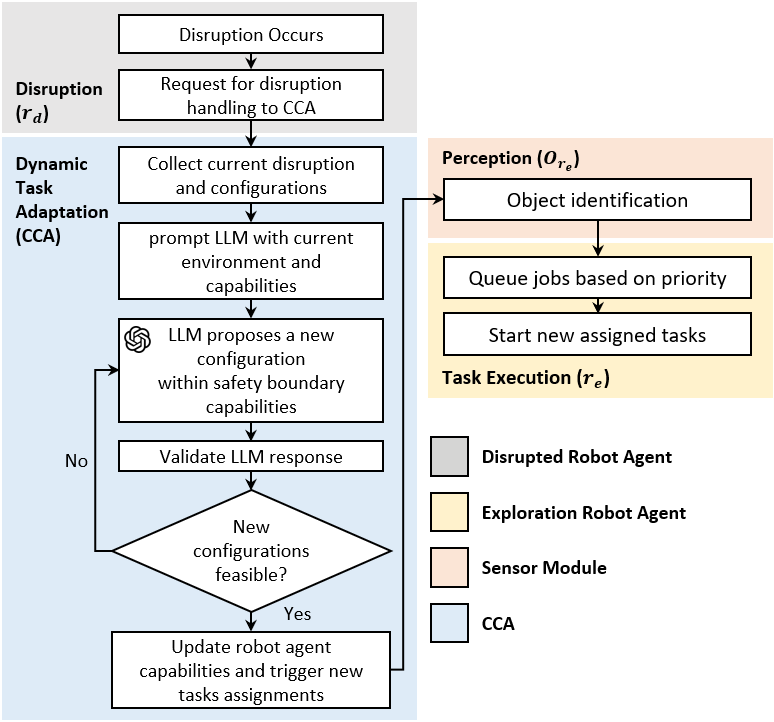}
    \caption{Implementation of dynamic task adaptation process}
    \label{fig:process}
\end{figure}

\begin{figure}[t]
\smallskip
\smallskip
    \captionsetup{belowskip=5pt}
    \includegraphics[width=.48\textwidth]{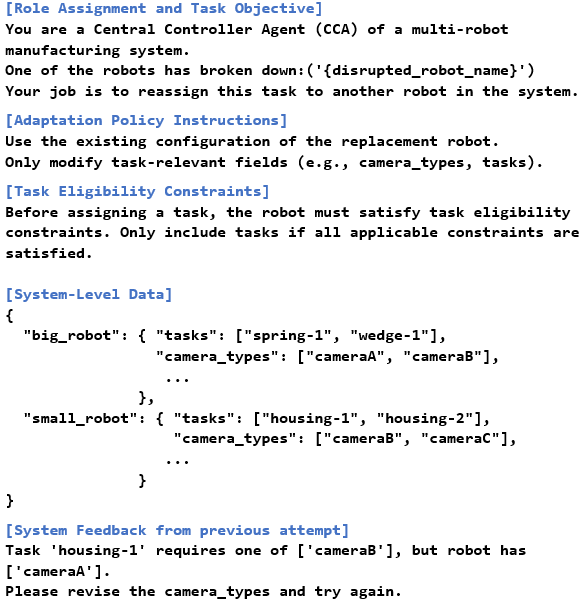}
    \caption{Simplified prompt provided to the \gls{llm} includes role assignment, adaptation policy instructions, task eligibility constraints, system-level configurations, and feedback from failed validation attempts}
    \label{fig:llmPrompt}
\end{figure}

\subsubsection{Dynamic Task Adaptation Process}
\label{subsec:process}

The process diagram, illustrated in Figure~\ref{fig:process}, shows the closed-loop dynamic task adaptation process initiated in response to a robot disruption. 

Upon failure detection by the disrupted robot $r_d$, the~\gls{cca} is notified and initiates a reassignment process. The~\gls{cca} gathers the current system context, such as disrupted robot configuration \( \mathcal{K}_{r_d} \), the set of alternative robots \( \mathcal{R} \), buffers, and available sensors. The collected context is input into a structured prompt and sent to the~\gls{llm}. The~\gls{llm} generates an updated configuration \( \mathcal{K}_{r_e}' \) for a selected candidate robot \( r_e \in \mathcal{R} \), referred to as the exploration robot. The updated configuration \( \mathcal{K}_{r_e}' \) is verified by the~\gls{cca} to ensure that the task \( \tau_d \) is valid under the proposed configuration. Specifically, the~\gls{cca} checks that all constraints in \( \mathcal{C}_{\tau_d} \) are satisfied. If the response fails validation, the~\gls{cca} provides structured feedback and re-prompts the~\gls{llm} until a feasible configuration is obtained or the maximum retry limit is reached. Once a valid reassignment is identified, the~\gls{cca} updates the configuration of \( r_e \), assigns the new task, and triggers task execution. The sensor module provides real-time perceptual data \(\mathcal{O}_{r_e} \), which the exploration robot $r_e$ uses to identify relevant objects originally assigned to the disrupted robots $r_e$ and execute the reassigned task accordingly. Tasks that violate the constraints are excluded from the assignment.

\subsubsection{Prompt Design for \gls{llm} Decision-Making}
\label{subsec:prompt}

The~\gls{cca} utilizes structured prompts to guide the \gls{llm} in generating valid dynamic task adaption in the event of a robot disruption. As illustrated in Figure~\ref{fig:llmPrompt}, the prompt is composed to inlcude both high-level task objectives and system-level constraints. The prompt contains the following components:

\begin{itemize}
    \item \textbf{Role Assignment and Task Objectives:} Instructs the \gls{llm} to understand the role of the~\gls{cca} and reassign a disrupted task \( \tau_d \in \mathcal{T} \) to an exploration robot \( r_e \in \mathcal{R} \).
    \item \textbf{Adaptation Policy Instructions:} Directs the \gls{llm} to use the existing configuration of the candidate robot \( \mathcal{K}_{r_e} \) as a baseline and update fields directly relevant to tasks.
    \item \textbf{Task Eligibility Constraints:} Instructs the~\gls{llm} to determine whether a task can be reassigned by reasoning over contextual conditions such as reachability, and sensor coverage inferred from configuration data.
    \item \textbf{System-Level Data:} Includes the full manufacturing environment configurations, including robots, buffers, and sensors.
    \item \textbf{Validation Feedback:} If a prior \gls{llm} response fails validation (e.g., assigning a task that violates spatial constraints), a structured error message is appended to the next prompt. This feedback loop enables iterative correction to a valid configuration.
\end{itemize}

This structured prompt design enables the \gls{llm} to reason over constraints, preserving safety boundaries, ensuring that task reassignments are feasible.

\section{Experimental Study}
\label{sec:experiments}

\subsection{Experimental Setup}
\label{subsec:experimental-setup}

\begin{figure}[t]
\smallskip
\smallskip
    \captionsetup{belowskip=-1pt}
    \includegraphics[width=.48\textwidth]{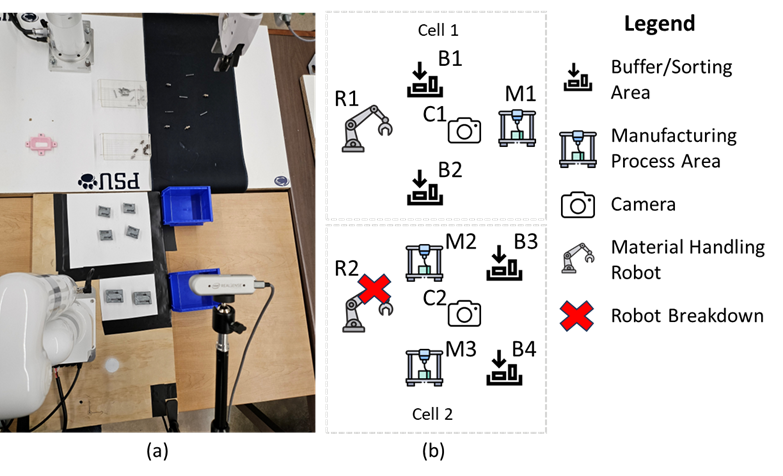}
    \caption{(a) Physical implementation of a multi-robot manufacturing system with real-time camera sensing and robotic arms handling parts. (b) Logical abstraction of the system divided into two cells, illustrating robot-task allocation and a simulated disruption in Cell 2 requiring dynamic adaptation}
    \label{fig:experimental-setup}
\end{figure}

To evaluate the proposed \gls{llm}-enabled dynamic task adaptation system, a physical multi-robot manufacturing environment was implemented, as shown in Figure~\ref{fig:experimental-setup}. The setup includes two robotic manipulators from UFactory~\cite{UFACTORY}, an xArm6 (\( R_1 \)) and a Lite6 (\( R_2 \)), distributed across two adjacent work cells. Each robot is equipped with object detection capabilities using both mounted and tripod-based cameras. The workspace includes buffer areas and processing stations for part-handling tasks.

Figure~\ref{fig:experimental-setup}(a) shows the real-world deployment, while (b) illustrates the system layout. Cell~1 is fully operational with robot \( R_1 \), camera \( C_1 \), buffers \( B_1 \), \( B_2 \), and processing machines \( M_1 \). Cell~2 simulates a disruption event where robot \( R_2 \) is considered non-functional (marked by a red X) after completing two part-sorting tasks, making it unavailable for tasks involving machines \( M_2 \), \( M_3 \) and buffers \( B_3 \), \( B_4 \).

The disrupted robot prompts the~\gls{cca} to enable the \gls{llm} for dynamic task reassignment. The system evaluates whether robot \( R_1 \) can take over tasks originally assigned to \( R_2 \) based on constraints such as spatial reachability, sensor coverage, and tool compatibility. Specifically, the reachability constraint of \( R_1 \) limits its access only up to \( M_2 \) and \( B_3 \), making it physically incapable of operating on \( M_3 \) or \( B_4 \).

The key objective of this experiment is to assess whether the \gls{llm} can infer these limitations from structured configuration data without being explicitly told which tasks are infeasible. This tests the \gls{llm}'s ability to reason over context—selecting only valid tasks (e.g., \( M_2 \), \( B_3 \))—based on implicit system constraints. Such capability demonstrates the potential of \gls{llm}-enabled systems to scale to more complex, real-world manufacturing scenarios where pre-programming of every possible failure and adaptation is impractical. In this experiment, we utilized GPT-4o from OpenAI~\cite{hurst2024gpt}.

\subsection{Experimental Results}
\label{subsec:experimental-results}

To evaluate the effectiveness of the proposed framework, a series of 20 trials were conducted simulating the failure of robot \( R_2 \), with the goal of reassigning a subset of its tasks to robot \( R_1 \). These trials assess the~\gls{llm}'s ability to generate valid configuration updates under real-world physical constraints, as well as the number of retry attempts required to obtain a successful reassignment. Additionally, the causes of unsuccessful adaptations were analyzed, and the average adaptation time was measured to evaluate real-time system responsiveness.

As summarized in Table~\ref{tab:summary_results}, the system achieved a 100\% success rate across all trials, meaning valid reassignment configurations were eventually found for every disruption. In 60\% of the cases (12 out of 20), the~\gls{llm} generated a valid reassignment on the first attempt. The remaining 40\% required iterative retries, supported by user-defined structured feedback from the~\gls{cca}, with a maximum of 4 retries.

The~\gls{llm} demonstrated an ability to interpret system-level constraints from configuration data without explicit instructions for every constraint. However, errors, such as assigning unreachable task locations still occurred. The validation and feedback process provided by the~\gls{cca} improved the quality of responses across retries. In safety-critical applications, validation criteria defined by the~\gls{sme} remain important to ensure the reliability of~\gls{llm}-enabled reconfigurations.

\begin{table}[t]
\centering
\caption{Summary of Dynamic Task Adaption Experiments across 20 trials}
\label{tab:summary_results}
\small
\begin{tabular}{|l|c|}
\hline
\textbf{Metric} & \textbf{Value} \\
\hline
Successful Reassignments & 20 \\
Success Rate & 100\% \\
Successful Reassignments (1st attempt) & 12 \\
Success Rate from the (1st attempt) & 60\% \\
Avg. Adaptation Time (success) & 19.36 s \\
Max Adaptation Time & 8.73 s \\
Min Adaptation Time & 34.9 s \\
Avg. LLM Retries (success) & 1.6 \\
Max LLM Retries & 4 \\
\hline
\end{tabular}
\end{table}

\subsection{Discussions from Experiments}
\label{subsec:discussions}

The experiments demonstrate the practical feasibility of using a~\gls{llm} in~\gls{cca} framework for dynamic task adaptation in a real-case scenario. In most trials, the system successfully generated valid task reassignment configurations within a few~\gls{llm} prompting attempts. These outcomes highlight the~\gls{llm}’s ability to reason over structured configuration data and adapt robotic behavior without relying on hardcoded rules.

However, the results also highlight limitations and areas for improvement. Failure cases often resulted in assigning unreachable dropoff zones, which indicate reasoning errors or hallucinations. Studies have shown that~\glspl{llm} struggle with spatial reasoning~\cite{zhang2025mitigating} and often fail to reliably apply constraints when they are embedded within complex or lengthy prompts~\cite{an2024make}. While the feedback mechanism can improve success rates over multiple retries, explicitly defining all possible constraint domains is impractical. Therefore, improving the robustness of~\gls{llm} reasoning under uncertain conditions remains a key research direction. Addressing this requires integrating structured domain knowledge base~\cite{lim2023ontology} to improve decision accuracy.

\section{Conclusion and Future Work}
\label{sec:conclusion}

This paper explored an \gls{llm}-enabled~\gls{cca} framework for dynamic task adaptation in multi-robot manufacturing systems. The proposed approach aims to support the system to respond to robot disruptions by generating and validating new task assignments based on structured configuration data. Unlike traditional rule-based approaches, this framework investigates the potential of reasoning capabilities of~\glspl{llm} to infer task feasibility under constraints.

Future work will focus on extending this approach to support more complex, lower-level operations such as adaptive assembly or multi-step manipulation tasks. This requires enhancing the~\gls{llm}'s reasoning with ability under uncertainty, particularly in scenarios involving multiple failures and dynamic environments. To address this, we plan to integrate runtime knowledge bases into the~\gls{cca} framework to provide contextual grounding and improve the consistency and safety of~\gls{llm}-generated outputs.



\balance


\end{document}